\documentclass[11pt]{article}
\usepackage[preprint]{acl}
\usepackage{times}
\usepackage{latexsym}
\usepackage[T1]{fontenc}
\usepackage[utf8]{inputenc}
\usepackage{microtype}
\usepackage{amsmath,amsfonts,amssymb}
\usepackage{array}
\usepackage{booktabs}
\usepackage{graphicx}
\usepackage{tikz}
\usetikzlibrary{arrows.meta,positioning,fit}
\usepackage{xspace}
\usepackage{enumitem}

\newcolumntype{P}[1]{>{\raggedright\arraybackslash}p{#1}}
\newcommand{\tablestyle}{%
  \fontsize{10}{12}\selectfont
  \setlength{\tabcolsep}{4pt}%
  \renewcommand{\arraystretch}{1.06}%
}
\newcommand{\tablehead}[1]{\multicolumn{1}{c}{#1}}
\newcommand{\method}{\textsc{Affirm}\xspace}
\newcommand{\directmethod}{\textsc{Affirm-Direct}\xspace}
\newcommand{\expertmethod}{\textsc{Affirm-Expert}\xspace}
\newcommand{\factorizedmethod}{\textsc{Affirm-K1}\xspace}
\newcommand{\globalfield}{\textsc{Direct Parent}\xspace}
\newcommand{\routefield}{\textsc{Route Field}\xspace}
\newcommand{\expertparent}{\textsc{Expert Parent}\xspace}

\title{Repairing Speech Access After Text Updates\\with Frozen Affine Fields}
\author{
  Ming-Hao Hsu \quad Yuxiang Wang \quad Yingda Shen \quad Zhizheng Wu \\
  School of Data Science \\
  The Chinese University of Hong Kong, Shenzhen, China \\
  \texttt{wuzhizheng@cuhk.edu.cn}
}
\hypersetup{
  pdftitle={Repairing Speech Access After Text Updates with Frozen Affine Fields},
  pdfauthor={Ming-Hao Hsu, Yuxiang Wang, Yingda Shen, Zhizheng Wu}
}

\begin{document}
\maketitle
\begin{abstract}
Speech-capable language models inherit most of their knowledge from text-rich language backbones, and continued model maintenance is likewise dominated by text updates.
Knowledge acquired through a text update can nevertheless remain difficult to access when the same query is spoken.
We formulate this mismatch as \emph{post-update speech access repair}, where the knowledge-bearing checkpoint stays frozen and a detachable operator repairs speech-only access in one model pass.
We introduce Affine Fields for Interface Repair across Modalities (\method), which first learns and freezes a base correction for speech hidden states, then adds query-specific corrections with zero mean over the operator-training queries.
Paired speech and text states provide the coordinates that condition these corrections on each query.
On balanced synthetic-speech PopQA, paired comparisons do not establish a clear advantage for \method over global SALSA steering or low-rank ReFT interventions under the same frozen single-state constraint.
On recorded SLUE-SQA-5 questions with text passages, a full-scale PopQA-fitted \method operator transfers without refitting and improves ROUGE-L by 0.0941 over its paired No Repair baseline.
Low-rank ReFT interventions also transfer positively, whereas the matched SALSA variant falls 0.0707 below its paired baseline.
These results show that text acquisition and speech access can diverge after an update and should be verified and repaired separately.
\end{abstract}

\section{Introduction}
\label{sec:intro}

Large language models increasingly serve as the reasoning core of interactive systems that use tools and act on external environments \citep{yao2023react,schick2023toolformer}.
Recent speech-language models extend this interaction from typed text to spoken input and output \citep{tang2024salmonn,hu2024wavllm,xu2025qwen25omni,shih2024gsqa}.
Most general-purpose training data and factual maintenance material for these systems still come from text: web pages, documents, code, and manuals are available at far larger scale than paired speech supervision \citep{raffel2020exploring,gao2020pile,wang2023blsp}.
Speech-capable models therefore commonly attach audio encoders and alignment modules to a text-pretrained language backbone, while text remains the direct route for factual adaptation and targeted editing \citep{tang2024salmonn,hu2024wavllm,xu2025qwen25omni,meng2022locating,meng2023memit}.

\begin{figure*}[!t]
\centering
\begingroup
\definecolor{affirmInk}{HTML}{1F2A33}
\definecolor{affirmMuted}{HTML}{65717B}
\definecolor{affirmLine}{HTML}{A7B1B9}
\definecolor{affirmCell}{HTML}{D3D9DE}
\definecolor{affirmPanel}{HTML}{F4F5F6}
\definecolor{affirmText}{HTML}{3D6DA8}
\definecolor{affirmSpeech}{HTML}{D18222}
\definecolor{affirmSpeechInk}{HTML}{A9640F}
\definecolor{affirmTeal}{HTML}{1A7F78}
\definecolor{affirmTealTint}{HTML}{E4F2F0}
\definecolor{affirmFail}{HTML}{C34A36}
\begin{tikzpicture}[
  x=1cm,y=1cm,
  font=\sffamily\fontsize{8}{9.5}\selectfont,
  text=affirmInk,
  line cap=round,line join=round,
  flow/.style={draw=affirmMuted,line width=.6pt,-{Stealth[length=1.5mm,width=1.15mm]}},
  repair/.style={flow,draw=affirmTeal},
  card/.style={draw=affirmLine,fill=white,line width=.5pt,rounded corners=2.5pt},
  chip/.style={fill=white,line width=.55pt,rounded corners=2pt},
  label/.style={inner sep=0pt,align=center},
  note/.style={label,text=affirmMuted,font=\sffamily\fontsize{7}{8.4}\selectfont},
  heading/.style={label,anchor=base west,font=\sffamily\bfseries\fontsize{9}{10.5}\selectfont},
  term/.style={label,anchor=base west,text=affirmTeal,
               font=\sffamily\bfseries\fontsize{7.5}{9}\selectfont},
  state/.style={circle,inner sep=0pt,minimum size=1.7mm,fill=affirmCell},
  waveform/.pic={
    \foreach \h [count=\i] in {.05,.10,.18,.26,.20,.12,.06,.09,.19,.30,.24,.28,
                               .16,.08,.05,.11,.22,.17,.25,.13,.07,.04}{
      \draw[affirmSpeech,line width=.95pt] (\i*.075-.8625,-\h)--(\i*.075-.8625,\h);
    }
  },
  snowflake/.pic={
    \foreach \a in {0,60,...,300}{
      \draw[affirmMuted,line width=.4pt,rotate=\a]
        (0,0)--(0,.105) (0,.058)--(-.03,.088) (0,.058)--(.03,.088);
    }
  },
  correct/.pic={
    \fill[affirmTeal] (0,0) circle (.125);
    \draw[white,line width=.8pt] (-.062,-.004)--(-.02,-.048)--(.064,.05);
  },
  wrong/.pic={
    \fill[affirmFail] (0,0) circle (.125);
    \draw[white,line width=.8pt] (-.046,-.046)--(.046,.046) (-.046,.046)--(.046,-.046);
  },
  query/.style={circle,inner sep=0pt,minimum size=1.2mm,fill=affirmSpeech},
  nudge/.style={draw=affirmTeal,line width=.6pt,-{Stealth[length=1.1mm,width=.9mm]},
                shorten <=.85mm}
]
\path[use as bounding box] (0,0) rectangle (16,5.32);

\fill[affirmPanel,rounded corners=3.5pt] (0,0) rectangle (5.97,4.92);
\fill[affirmPanel,rounded corners=3.5pt] (6.20,0) rectangle (16,4.92);
\node[heading] at (.02,5.08) {(a)\enspace After a text-only update};
\node[heading] at (6.22,5.08) {(b)\enspace \textsc{Affirm} repairs one hidden state};

\path[card] (1.30,1.5) rectangle (5.80,3.7);
\draw[affirmText!55,line width=.5pt] (2.40,1.5)--(2.40,3.7);
\draw[affirmSpeech!60,line width=.5pt] (4.70,1.5)--(4.70,3.7);
\foreach \y in {1.82,2.21,2.60,2.99,3.38}{
  \node[state,fill=affirmText!55] at (2.40,\y) {};
  \node[state,fill=affirmSpeech!60] at (4.70,\y) {};
}
\node[label] at (3.55,2.80) {acquired\\checkpoint};
\node[label,font=\fontsize{9}{10}\selectfont] at (3.55,2.24) {$F_{\theta^{+}}$};

\begin{scope}[shift={(.60,2.95)}]
  \draw[affirmText,fill=white,line width=.5pt]
    (-.17,-.22)--(-.17,.22)--(.07,.22)--(.17,.12)--(.17,-.22)--cycle;
  \draw[affirmText,line width=.5pt] (.07,.22)--(.07,.12)--(.17,.12);
  \foreach \y in {.03,-.05,-.13}{\draw[affirmText,line width=.45pt] (-.10,\y)--(.10,\y);}
\end{scope}
\draw[flow,draw=affirmText] (.87,2.95)--(1.30,2.95);
\node[note,text=affirmText] at (.63,2.40) {text-only\\update};

\path[chip,draw=affirmText] (1.30,.5) rectangle (3.50,1.2);
\node[label,font=\sffamily\fontsize{7}{8.6}\selectfont] at (2.40,.85)
  {What is the capital\\of Kansas?};
\path[chip,draw=affirmSpeech] (3.60,.5) rectangle (5.80,1.2);
\pic at (4.70,.85) {waveform};
\node[note] at (3.55,.26) {the same question, typed and spoken};
\draw[flow] (2.40,1.2)--(2.40,1.5);
\draw[flow] (4.70,1.2)--(4.70,1.5);

\draw[flow] (2.40,3.7)--(2.40,4.0);
\draw[flow] (4.70,3.7)--(4.70,4.0);
\node[label,anchor=base] (ansT) at (2.24,4.13) {Topeka};
\pic at ([xshift=2.1mm,yshift=.95mm]ansT.base east) {correct};
\node[label,anchor=base] (ansS) at (4.54,4.13) {Manhattan};
\pic at ([xshift=2.1mm,yshift=.95mm]ansS.base east) {wrong};
\draw[affirmFail,line width=.5pt] (2.40,4.43)--(2.40,4.51)--(4.70,4.51)--(4.70,4.43);
\node[note,text=affirmFail,anchor=base] at (3.55,4.60) {speech-access gap};

\path[card] (6.30,1.5) rectangle (9.36,3.7);
\fill[affirmTealTint,rounded corners=2pt] (6.38,2.43) rectangle (9.28,2.77);
\foreach \x in {7.52,7.90,8.28,8.66}{
  \draw[affirmCell,line width=.4pt] (\x,1.5)--(\x,3.7);
  \foreach \y in {1.82,2.21,2.60,2.99,3.38}{\node[state] at (\x,\y) {};}
}
\draw[affirmSpeech!60,line width=.5pt] (9.06,1.5)--(9.06,2.60);
\draw[affirmTeal!55,line width=.5pt] (9.06,2.60)--(9.06,3.7);
\foreach \y in {1.82,2.21}{\node[state,fill=affirmSpeech!60] at (9.06,\y) {};}
\foreach \y in {2.99,3.38}{\node[state,fill=affirmTeal!50] at (9.06,\y) {};}
\node[state,minimum size=2.6mm,fill=affirmTeal,draw=white,line width=.6pt] (edit)
  at (9.06,2.60) {};
\node[note,anchor=base east] at (7.30,1.74) {block $1$};
\node[note,anchor=base east,text=affirmInk] at (7.30,2.52) {block $\ell$};
\node[note,anchor=base east] at (7.30,3.30) {block $L$};
\pic[scale=1.2] at (6.43,3.92) {snowflake};
\node[note,anchor=base west] at (6.61,3.84) {frozen $F_{\theta^{+}}$};

\path[chip,draw=affirmSpeech] (7.29,.5) rectangle (8.89,1.2);
\pic[xscale=.82] at (8.09,.85) {waveform};
\node[note,text=affirmSpeechInk] at (8.09,.26) {spoken only, no transcript};
\draw[flow] (8.09,1.2)--(8.09,1.5);
\draw[affirmMuted,line width=.45pt] (9.06,1.5)--(9.06,1.30);
\node[note,anchor=north west,align=left] at (9.02,1.24) {final prompt\\position};
\draw[repair] (9.06,3.7)--(9.06,4.0);
\node[label,anchor=base] (ansR) at (8.90,4.13) {Topeka};
\pic at ([xshift=2.1mm,yshift=.95mm]ansR.base east) {correct};

\draw[repair] (edit.48) to[out=40,in=180] (9.58,2.80)--(11.05,2.80);
\draw[repair] (11.05,2.40)--(9.58,2.40) to[out=180,in=-40] (edit.-48);
\node[note,anchor=south,text=affirmInk] at (10.24,2.86) {$h^{S}$};
\node[note,anchor=north,text=affirmTeal] at (10.22,2.33) {$+\,\alpha R_\phi(h^{S})$};

\path[draw=affirmTeal,fill=affirmTealTint,line width=.65pt,rounded corners=2.5pt]
  (11.05,1.35) rectangle (15.85,3.85);
\node[label,anchor=base west,text=affirmTeal,
      font=\sffamily\bfseries\fontsize{8}{9.5}\selectfont] at (11.05,4.47)
  {\textsc{Affirm} operator $R_\phi$};
\node[note,anchor=base west] at (11.05,4.13) {detachable, fitted on typed--spoken pairs};

\path[fill=white,rounded corners=1.5pt] (11.20,2.80) rectangle (12.65,3.71);
\path[fill=white,rounded corners=1.5pt] (11.20,1.49) rectangle (12.65,2.40);
\foreach \x/\y/\dx/\dy in {.69/.20/.27/.07,.99/.62/-.18/-.20,.33/.20/.23/-.10,
                           .65/.65/-.29/-.05,.14/.54/-.03/.28}{
  \node[query] at (11.26+\x,2.80+\y) {};
  \draw[nudge] (11.26+\x,2.80+\y)--++(.28,.15);
  \node[query] at (11.26+\x,1.49+\y) {};
  \draw[nudge] (11.26+\x,1.49+\y)--++(\dx,\dy);
}
\node[term] at (12.85,3.495) {shared access parent};
\node[label,anchor=base west] at (12.85,3.185) {$G_\psi$};
\node[note,anchor=base west] at (12.85,2.895) {shared across queries};
\draw[affirmTeal,line width=.8pt] (11.845,2.60)--(12.005,2.60) (11.925,2.52)--(11.925,2.68);
\node[term] at (12.85,2.185) {centered route field};
\node[label,anchor=base west] at (12.85,1.875) {$\beta\,U\bigl(z(h^{S})-\bar z\bigr)$};
\node[note,anchor=base west] at (12.85,1.585) {query-specific, zero mean};

\node[query] at (11.13,.88) {};
\node[note,anchor=base west] at (11.26,.80) {$h^{S}$ of one question};
\draw[nudge,shorten <=0pt] (13.82,.88)--(14.12,.88);
\node[note,anchor=base west] at (14.22,.80) {added shift};
\end{tikzpicture}
\endgroup
\caption{\textbf{Decoupled knowledge acquisition and speech access repair.}
(a) Text-only acquisition can leave a speech-access gap.
(b) \method combines a base correction (shared access parent) with a query-specific correction (centered route field) and edits the final prompt state with the acquired checkpoint frozen.
The example shows recorded outputs for one PopQA test question held out from operator fitting, with \directmethod as the repair.}
\label{fig:overview}
\end{figure*}
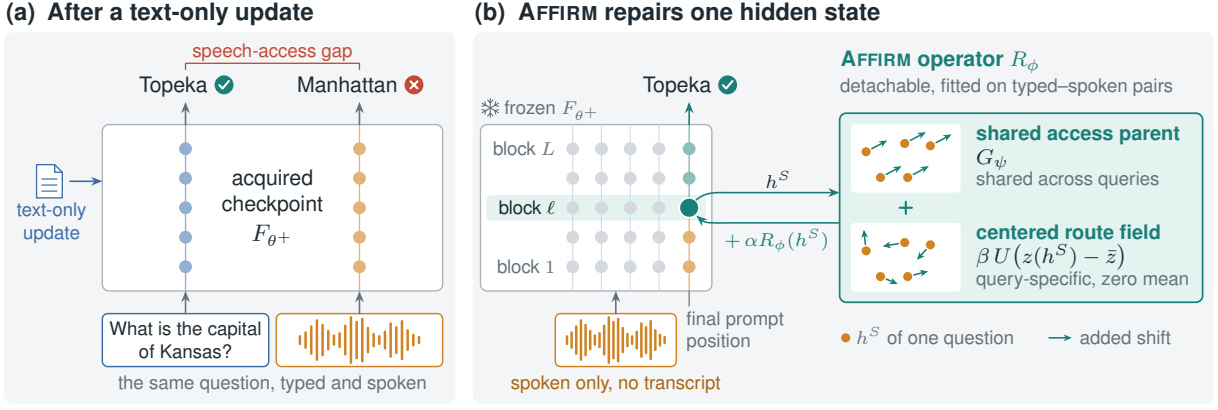

Native multimodality does not by itself make every text update equally usable from speech.
Studies of speech-language models report differences between matched spoken and typed inputs in semantic understanding, reasoning, and factual-recall pathways \citep{xiang2025modalitygap,cuervo2026closing,modica2026factual,hsu2026geometric,hsu2026entitybinding}.
We study a post-update instance of this mismatch: after a successful text-only update, a model may answer an updated fact when the question is typed yet fail when the same question is spoken.
The correct text response shows that the acquired checkpoint can express the updated fact, while the paired speech failure defines the post-update speech-access gap that we repair.

This problem arises whenever knowledge is maintained through text while users continue to query the resulting system by voice.
We consider a contract in which the acquired checkpoint is immutable and the repair remains detachable.
At deployment, the question is supplied only as speech, and the repair edits one decoder state in one model pass without a transcript.
The contract separates the knowledge-bearing checkpoint from the interface that makes its updated knowledge accessible (Figure~\ref{fig:overview}).

The central design question is how to combine a base repair learned across speech queries with a correction tailored to each input.
We combine these roles in \method, or \emph{Affine Fields for Interface Repair across Modalities}.
We first learn and freeze a base correction, called the \emph{shared access parent}.
We then add a query-specific correction, called the \emph{centered route field}, whose mean over the operator-training queries is zero.
Paired spoken and typed queries provide the hidden-state coordinates used to condition this correction.

The experiments first establish a large paired text--speech gap after successful text acquisition.
On balanced PopQA, paired bootstrap comparisons do not establish a clear advantage for \directmethod over DiReFT, LoReFT, or SALSA.
When operators fitted on full-scale PopQA transfer without refitting to recorded SLUE questions, \directmethod and ReFT improve over No Repair, whereas the matched SALSA variant degrades.
Parent comparisons and a two-seed depth sweep examine which design choices affect performance, and the compact variant also improves held-out access on Kimi-Audio.

Our contributions are:
\begin{itemize}[leftmargin=1.35em,itemsep=2pt,topsep=3pt]
    \item We formulate post-update speech access repair, separating text knowledge acquisition from detachable speech access through an immutable acquired checkpoint.
    \item We introduce \method, a staged affine field that combines a shared access parent with a centered query-specific route field learned from paired speech--text routes.
    \item We show that methods with comparable synthetic-speech performance can differ under recorded-speech transfer, where \method and ReFT improve over No Repair while the tested SALSA variant regresses.
\end{itemize}

\section{Related Work}
\label{sec:related}

\paragraph{Text-centered speech language models and modality gaps.}
Modern speech language models commonly connect an audio encoder to a pretrained text decoder through projectors, adapters, and staged multimodal training \citep{tang2024salmonn,hu2024wavllm,xu2025qwen25omni,hsu2026cgate}.
Behavior alignment and cross-modal distillation train the speech route to reproduce capabilities available from text \citep{wang2023blsp,wang2024blspkd,wang2025crossmodalkd,hu2026cord,cao2026xopd}.
Speech inputs nevertheless remain weaker on some semantic and reasoning tasks, and representation similarity is associated with the remaining modality gap \citep{xiang2025modalitygap,cuervo2026closing,wang2026tars}.
These studies improve speech during multimodal adaptation, whereas we address the later maintenance stage after a text update has established the target knowledge.

\paragraph{Cross-modal knowledge acquisition and access.}
Multimodal models can store or acquire information through one route without making it equally available from another \citep{wen2025crossmodal,basu2024information}.
Speech and text can also rely on distinct factual-recall pathways inside a shared multimodal model \citep{modica2026factual}.
Knowledge-editing methods change stored associations in language and audio-language models \citep{meng2022locating,meng2023memit,chung2026localizing}.
Our task instead begins after acquisition succeeds and repairs the route by which a frozen checkpoint accesses the acquired fact.

\paragraph{Frozen representation interventions.}
A complementary line of work leaves pretrained model weights fixed and modifies behavior through narrow interventions on hidden representations.
Inference-Time Intervention learns directions at selected attention heads, while activation addition modifies residual-stream states with behavior-specific directions \citep{li2023inference,turner2024steering}.
ReFT generalizes learnable representation interventions at chosen layers and positions, including the DiReFT and LoReFT parameterizations \citep{wu2024reft}.
These methods adapt behavior through a fixed text-model backbone, but they do not separate a text-acquisition stage from speech access to the acquired knowledge.
\method brings this intervention family to post-update speech access by learning from paired speech--text routes and retaining a detachable repair around a frozen shared parent.

\paragraph{Global and conditional steering.}
Learned speech steering can optimize shared layer-wise additive directions with a frozen backbone \citep{yegorova2026salsa}.
Input-dependent steering predicts an intervention from each multimodal context \citep{parekh2025learning}, while Steer-MoE dynamically mixes expert directions inside audio-encoder layers \citep{feng2025steermoe}.
Angle and length interventions provide another representation-level route for reducing the speech--text modality gap \citep{xiang2025modalitygap}.
Together, these approaches establish global and input-conditioned steering as complementary mechanisms for reducing speech--text discrepancies.
\method addresses post-update factual access through a detachable operator that edits one decoder state of a frozen acquired checkpoint.

\section{Preliminaries and Problem Formulation}
\label{sec:problem}

\subsection{Post-update access}

Let $F_{\theta^+}$ be a speech-capable autoregressive checkpoint obtained by text-only answer-likelihood training from $F_{\theta_0}$.
Each fact $i$ has a reference answer $y_i=(y_{i,1},\ldots,y_{i,T_i})$ and semantically matched text and speech queries $x_i^T$ and $x_i^S$.
For modality $m\in\{T,S\}$ at decoder block $\ell$, let $h_i^m\in\mathbb R^d$ denote the hidden state at the final attended prompt position before the model predicts $y_{i,1}$.
The two states are directly comparable because they come from the same checkpoint and correspond to matched queries.
We verify $F_{\theta^+}$ through the text route and then freeze it.
For acquired-fact evaluation on PopQA and TOFU, the disjoint operator-training, development, and evaluation sets $\mathcal D_{\mathrm{op}}$, $\mathcal D_{\mathrm{dev}}$, and $\mathcal D_{\mathrm{eval}}$ are subsets of the text-acquisition set.
Evaluation facts provide no operator gradients or selection signals.

For any scored set $\mathcal E$ and scoring function $q$, route-specific access is
\begin{equation}
\begin{aligned}
    \operatorname{Access}_{m}(F_{\theta^+};\mathcal E)
    &=\frac{1}{|\mathcal E|}
    \sum_{i\in\mathcal E}
    q\!\left(F_{\theta^+}(x_i^m),y_i\right), \\
    \Delta_{T\rightarrow S}
    &=\operatorname{Access}_{T}-\operatorname{Access}_{S}.
\end{aligned}
    \label{eq:access_gap}
\end{equation}
A positive $\Delta_{T\rightarrow S}$ is an operational post-update access gap: the acquired checkpoint can answer through text more reliably than through the matched speech interface.

\subsection{Frozen repair contract}

After text acquisition, we fit a detachable repair operator $R_\phi:\mathbb R^d\rightarrow\mathbb R^d$ on $\mathcal D_{\mathrm{op}}$ and deploy it as
\begin{equation}
\begin{aligned}
    \widetilde h_i^S &= h_i^S + \alpha R_\phi(h_i^S), \\
    \widehat y_i &= F_{\theta^+}\!\left(
        x_i^S;\,h_i^S\leftarrow\widetilde h_i^S
    \right),
\end{aligned}
    \label{eq:repair_contract}
\end{equation}
where $\theta^+$ remains frozen during operator training and inference.
The scalar $\alpha$ controls intervention strength and is fixed before test evaluation.
The question is supplied only as speech, without a transcript, matched text-question state, retrieved example, or second model pass.
Text-query success establishes acquisition, the paired route gap diagnoses inaccessible behavior, and held-out speech improvement measures repair.
We separately test transfer to SLUE, where recorded questions accompany text passages and the task, prompt, speakers, and acoustics change together without refitting.

\section{\method: Shared Access Parent with Centered Route Field}
\label{sec:method}

\subsection{Method overview}

\method decomposes post-update repair into a shared access parent and a query-conditioned route field around that parent.
We first learn a low-dimensional code that predicts the difference between matched text and speech prompt states, without answer labels.
Stage A trains the parent using answer supervision through the frozen decoder.
Stage B freezes the parent and the coordinate map, then trains only the readout that converts these codes into query-specific corrections.
The acquired checkpoint remains frozen throughout, and deployment stores only the compact operator.

\subsection{Affine repair field}

For a speech state $h^S\in\mathbb R^d$ at decoder block $\ell$, every \method variant adds the parent and centered correction:
\begin{equation}
\begin{aligned}
    R_\phi(h^S)
    &=G_\psi(h^S)+\beta U\bigl(z(h^S)-\bar z\bigr), \\
    \widetilde h^S&=h^S+\alpha R_\phi(h^S).
\end{aligned}
    \label{eq:affine_field}
\end{equation}
Here $G_\psi$ is the shared access parent, $z(h^S)\in\mathbb R^r$ is a paired-route coordinate, and $U\in\mathbb R^{d\times r}$ maps route variation back to the decoder state.
The coefficients $\beta$ and $\alpha$ control the route-field strength and overall intervention strength, respectively.
With $\bar z=|\mathcal D_{\mathrm{op}}|^{-1}\sum_i z(h_i^S)$, the route field has zero mean on the operator-training population and therefore represents variation around the fitted parent.
The route field is affine in the route coordinate $z$, while $z$ itself depends nonlinearly on $h^S$ through Equation~\ref{eq:route_objective}.

\subsection{Answer-free paired-route coordinates}

The frozen checkpoint produces paired states $\{(h_i^S,h_i^T)\}_{i=1}^N$ at block $\ell$, where $N=|\mathcal D_{\mathrm{op}}|$ and $\Delta_i=h_i^T-h_i^S$.
Let $\mu_S$ and $\bar\Delta$ be the corresponding population means, and let $\sigma_S$ and $\sigma_\Delta$ be scalar root-mean-square scales over samples and dimensions.
We fit a low-rank coordinate map $D$ and temporary readout $V$ using
\begin{equation}
\begin{aligned}
    z_i&=\operatorname{SiLU}\!\left(D\frac{h_i^S-\mu_S}{\sigma_S}\right),
    \\
    \delta_i&=\frac{\Delta_i-\bar\Delta}{\sigma_\Delta}, \\
    D^*,V^*&=\arg\min_{D,V}\frac{1}{Nd}\sum_i
    \left\lVert Vz_i-\delta_i\right\rVert_2^2.
\end{aligned}
    \label{eq:route_objective}
\end{equation}
The fit uses paired prompt states but no answer labels.
We freeze the coordinate map $D^*$ and discard the temporary readout $V^*$, whose role was to predict state differences.
Stage B learns a separate readout $U$ to turn these fixed coordinates into corrections that improve answer generation.

\subsection{Stage A: decoder-aware access parents}

Stage A trains the parent through the frozen decoder using teacher-forced answer cross-entropy.
If $p_{i,t}^{G}$ denotes the answer distribution after applying the parent to $h_i^S$, the objective is
\begin{equation}
    \mathcal L_A(\psi)
    =-\frac{1}{N}\sum_i\frac{1}{T_i}
    \sum_{t=1}^{T_i}\log p_{i,t}^{G}(y_{i,t}),
    \label{eq:parent_objective}
\end{equation}
where prompt tokens are masked and only $\psi$ receives gradients.
The direct and expert parents are
\begin{equation}
\begin{aligned}
    G_{\mathrm{direct}}(h)&=g, \\
    G_{\mathrm{expert}}(h)&=s\sum_{k=1}^{K}\operatorname{softmax}(Wh)_k e_k,
\end{aligned}
    \label{eq:parent_parameterizations}
\end{equation}
where $g$ is one shared vector, $W$ is a router, $e_k$ are learned directions, and $s$ is a learned scale.
All parent parameters are optimized with Equation~\ref{eq:parent_objective} and frozen before Stage B.
In the balanced PopQA diagnostic, the $K=64$ softmax concentrates almost all weight on the same expert across operator-training states, yielding a nearly constant correction.
In this diagnostic, the expert bank acts as an alternative parameterization of a shared correction.
We also use the compact $K=1$ form $G(h)=se$ to study intervention depth and cross-backbone portability, with \directmethod as the primary variant.

\subsection{Stage B: selective correction around a frozen parent}
\label{sec:curriculum}

Stage B initializes $U=0$ and trains only $U$ while the checkpoint, $D^*$, and the selected parent remain frozen.
For each fitting item, teacher-forced decoding supplies text-route and parent-only speech logits.
We choose the text route when its mean gold-answer log probability exceeds the parent route by more than threshold $\tau$, and otherwise use the parent route as a preservation teacher.
Let $\pi_{i,t}^*$ be the selected teacher's top-$k$ distribution after temperature scaling and renormalization, and let $p_{i,t}^{U}$ be the full-field speech distribution.
The readout minimizes
\begin{equation}
    U^*=\arg\min_U
    \sum_i\frac{1}{NT_i}\sum_{t=1}^{T_i}
    D_{\mathrm{KL}}\bigl(
        \operatorname{sg}(\pi_{i,t}^*)
        \,\|\,
        p_{i,t}^{U}
    \bigr),
    \label{eq:stage_b_objective}
\end{equation}
where $\operatorname{sg}$ denotes stop-gradient.
The text teacher supplies correction when text retains a clear answer-likelihood advantage, while the parent teacher preserves an already useful answer distribution.
Zero initialization makes the first forward pass in Stage B reproduce the parent-only predictions, and backpropagation through the frozen decoder updates only $U$.

\subsection{Fitting and deployment}

For the primary Qwen experiments, we select the block, overall strength $\alpha$, and route-field coefficient $\beta$ on development data, then fix the operator before testing.
At inference, \method edits the final prompt state during prefill and then continues ordinary greedy generation without editing answer-token states.
It requires no question transcript, matched text-question state, pre-update checkpoint $F_{\theta_0}$, or external ASR system.
Appendix~\ref{app:reproduction} gives the fitting schedules and inference settings for each variant.

\section{Experimental Setup}
\label{sec:setup}

\paragraph{Model and factual acquisition.}
We use Qwen2.5-Omni-7B~\citep{xu2025qwen25omni}.
For full-scale experiments, we fine-tune one checkpoint on text question--answer pairs from all 14,267 PopQA facts over 16 relations \citep{mallen2023trust} and another on all 4,000 TOFU questions over 200 fictional authors \citep{maini2024tofu}.
We then freeze each checkpoint and fit its repair operators.
We also evaluate compact operators on separately acquired Kimi-Audio-7B \citep{kimiteam2025kimiaudio} and MiMo-Audio-7B \citep{coreteam2025mimoaudio} checkpoints (Appendix~\ref{app:cross_model}).

\paragraph{Full-scale operator splits.}
For PopQA, we group questions linked by a shared subject or normalized question string before splitting, so no group crosses the 8,572/2,848/2,847 operator train/development/test split.
PopQA-20 and PopQA-60 use nested subsets of 2,853 and 8,560 training examples, approximately 20\% and 60\% of the full acquisition corpus, with the same development and test sets.
TOFU uses 120/40/40 train/development/test authors, yielding 2,400/800/800 examples without author overlap.

\paragraph{Balanced comparison and baseline fitting.}
For the balanced PopQA comparison, we use a separate checkpoint acquired on 1,000 facts and a 600/199/201 operator train/development/test split that keeps duplicate questions together.
We separately fit each method on PopQA-20 for recorded-speech transfer and on TOFU training authors for author-disjoint evaluation.
Appendix~\ref{app:baseline_replication} gives the implementation and development-selection details for every reproduced baseline.

\paragraph{Speech-only baselines.}
The frozen-checkpoint comparison evaluates the following speech-only representation-level repair methods:
\begin{itemize}[leftmargin=1.35em,itemsep=2pt,topsep=3pt]
    \item \textbf{No Repair} applies the acquired checkpoint to speech without an added intervention.
    \item \textbf{DiReFT and LoReFT}~\citep{wu2024reft} learn low-rank hidden-state interventions at a selected decoder state.
    \item \textbf{L2S}~\citep{parekh2025learning} predicts an input-dependent steering intervention from the multimodal context.
    \item \textbf{SALSA}~\citep{yegorova2026salsa} learns one shared speech-aware steering activation vector with a frozen backbone.
\end{itemize}
All speech-only methods receive the question as audio, without its transcript or a paired text-question state, and use one pass through the frozen acquired checkpoint.
Speech LoRA~\citep{hu2022lora}, speech-to-text knowledge distillation \citep[S2T-KD,][]{wang2025crossmodalkd}, a Whisper ASR cascade~\citep{radford2023whisper}, and full speech supervision are reported separately as operational alternatives because they relax the frozen single-state deployment contract.

\paragraph{Donor-assisted controls.}
Angle and Length~\citep{xiang2025modalitygap} modify speech representations through direction and norm adjustments using a paired text-question state as a per-example inference-time donor.
We report them separately because the additional typed question is unavailable to speech-only methods.

\paragraph{Speech and transfer.}
CosyVoice renders the PopQA and TOFU questions used for fitting and evaluation \citep{du2024cosyvoice}.
Our transfer evaluation uses all 256 recorded SLUE-SQA-5 questions from 50 human speakers and five source datasets \citep{shon2023sluephase2}.
Each SLUE input contains a text passage and a spoken question, so speech-only refers to the question modality.
We apply each frozen PopQA-20 operator with no SLUE training examples, refitting, or target-domain strength retuning.
The separate Angle and Length controls access the paired text question at inference.
This tests transfer to passage-based question answering under changes in task, prompt, speakers, and acoustics, rather than recall of the acquired PopQA facts.

\paragraph{\method variants.}
Each variant combines a parent with the centered route field:
\begin{itemize}[leftmargin=1.35em,itemsep=2pt,topsep=3pt]
    \item \textbf{\directmethod} combines a direct shared parent with the centered route field and is the primary \method variant.
    \item \textbf{\expertmethod} combines the expert parent with the centered route field after Stage A.
    \item \textbf{\factorizedmethod} uses a compact scale--direction parent for depth and cross-backbone characterization.
\end{itemize}

\paragraph{Optimization, selection, and metrics.}
The hidden width is $d=3,584$ and low-rank operators use $r=64$.
Stage B freezes its parent and $D$, trains zero-initialized $U$, and deploys it with fixed $\beta=0.25$.
We sweep $\ell\in\{4,8,12,16,20,24,28\}$ only on development data, calibrate matched edit norms before test generation, and report ROUGE-L recall (RL), normalized exact match (EM), and SLUE answer-span accuracy.
Appendix~\ref{app:reproduction} gives complete schedules, doses, prompts, counts, seeds, and uncertainty procedures.

\section{Results}
\label{sec:results}

\subsection{The text update leaves a measurable speech-access gap}

Text-only acquisition succeeds without making the acquired knowledge equally usable from speech.
On CosyVoice-rendered paired questions over all 14,267 PopQA facts, the acquired checkpoint reaches 0.9213 RL and 0.9187 EM through text, compared with 0.1840 RL and 0.1519 EM through speech.
The held-out PopQA and TOFU questions test access to facts included in text acquisition.
To test whether speech states retain question-specific information, we retrieve each speech state's matching text state among 600 paired questions.
After subtracting each modality's mean state, top-1 retrieval accuracy is 50.3\%, compared with 0.17\% chance (Appendix~\ref{app:controls}).

\subsection{Methods comparable on synthetic speech diverge on recorded speech}

Table~\ref{tab:headline_results} compares methods on balanced PopQA, author-disjoint TOFU, and recorded SLUE questions.
Learned operators are fitted separately on balanced PopQA, TOFU, and PopQA-20, with only the PopQA-20 operator transferred to SLUE.

\begin{table*}[t]
\centering
\tablestyle
\begin{tabular*}{\linewidth}{@{\extracolsep{\fill}}P{4.7cm}rrrrrrr}
\toprule
& \multicolumn{2}{c}{Balanced PopQA} & \multicolumn{2}{c}{TOFU} & \multicolumn{3}{c}{SLUE no-refit} \\
\cmidrule(lr){2-3}\cmidrule(lr){4-5}\cmidrule(lr){6-8}
\tablehead{Method} & \tablehead{RL} & \tablehead{EM} & \tablehead{RL} & \tablehead{EM} & \tablehead{Span} & \tablehead{RL} & \tablehead{EM} \\
\midrule
No Repair & 0.2736 & 0.2189 & 0.5003 & 0.1000 & 0.4063 & 0.4435 & 0.2344 \\
DiReFT~\citep{wu2024reft} & 0.5945 & 0.5821 & 0.5117 & 0.1075 & 0.4609 & 0.4981 & 0.2734 \\
LoReFT~\citep{wu2024reft} & 0.5995 & 0.5920 & 0.5179 & 0.1138 & 0.4766 & 0.5209 & \textbf{0.3047} \\
L2S~\citep{parekh2025learning} & 0.5361 & 0.5274 & 0.5107 & 0.1000 & 0.3906 & 0.4261 & 0.2656 \\
SALSA~\citep{yegorova2026salsa} & 0.6119 & 0.6020 & 0.5237 & 0.1150 & 0.3203 & 0.3730 & 0.2656 \\
\midrule
\directmethod & \textbf{0.6318} & \textbf{0.6219} & \textbf{0.5310} & \textbf{0.1200} & \textbf{0.4922} & \textbf{0.5376} & 0.3008 \\
\bottomrule
\end{tabular*}
\caption{\textbf{Post-update factual access across source repair, author-disjoint evaluation, and no-refit recorded-speech transfer.}
PopQA reports the balanced test, TOFU reports 800 author-disjoint questions after TOFU fitting, and SLUE reports 256 recorded questions with text passages after PopQA-20 fitting without target-task training examples or strength retuning.
RL denotes ROUGE-L recall, EM denotes normalized exact match, and Span denotes SLUE answer-span accuracy.
All rows use one speech-only pass without a text-question donor, and \directmethod edits decoder block 20.
The displayed No Repair row is paired with \method and ReFT.
On SLUE, L2S and SALSA use their separately generated No Repair outputs for the paired differences reported in the text and Appendix Table~\ref{tab:slue_paired_baselines}.}
\label{tab:headline_results}
\end{table*}

\begin{table*}[t]
\centering
\tablestyle
\begin{tabular*}{\linewidth}{@{\extracolsep{\fill}}P{4.7cm}rrrrrrr}
\toprule
& \multicolumn{2}{c}{Balanced PopQA} & \multicolumn{2}{c}{TOFU} & \multicolumn{3}{c}{SLUE no-refit} \\
\cmidrule(lr){2-3}\cmidrule(lr){4-5}\cmidrule(lr){6-8}
\tablehead{Method} & \tablehead{RL} & \tablehead{EM} & \tablehead{RL} & \tablehead{EM} & \tablehead{Span} & \tablehead{RL} & \tablehead{EM} \\
\midrule
No Repair & 0.2769 & 0.2239 & 0.4996 & 0.0988 & 0.4102 & 0.4437 & \textbf{0.2422} \\
Angle~\citep{xiang2025modalitygap} & \textbf{0.5398} & \textbf{0.4925} & \textbf{0.6547} & \textbf{0.2550} & \textbf{0.4219} & \textbf{0.4619} & 0.2227 \\
Length~\citep{xiang2025modalitygap} & 0.2794 & 0.2239 & 0.5050 & 0.1025 & 0.4141 & 0.4476 & 0.2344 \\
\bottomrule
\end{tabular*}
\caption{\textbf{Donor-assisted controls under a relaxed inference condition.}
Angle and Length modify the speech state using the paired text-question state at inference.
No Repair reports the outputs generated alongside these controls on each dataset.
RL, EM, and Span follow the definitions in Table~\ref{tab:headline_results}.}
\label{tab:donor_controls}
\end{table*}

On balanced synthetic-speech PopQA, SALSA and \directmethod recover substantial access, and the paired-bootstrap comparisons with DiReFT, LoReFT, or SALSA do not resolve an advantage for \directmethod (Appendix Table~\ref{tab:published_intervals}).
Among the speech-only rows, \directmethod provides the strongest TOFU RL and EM point estimates.

\paragraph{Donor-assisted control.}
Supplying the paired text-question state gives the model access to the typed question at inference.
Under this relaxed condition, Angle improves TOFU RL and EM over No Repair, whereas Length remains near the no-repair point (Table~\ref{tab:donor_controls}).

On SLUE, \directmethod improves RL by 0.0941 and LoReFT improves by 0.0774, while SALSA falls 0.0707 below its paired No Repair baseline.
These operators use no SLUE training examples, question transcripts, or target-task strength selection.
The SLUE SALSA operator is fitted on PopQA-20, with its strength calibrated to match \method's mean edit norm on source speech states.
In a separate control, we recalibrate SALSA on unlabeled SLUE inputs to match \method's mean edit norm, and SALSA still falls below No Repair (Appendix~\ref{app:baseline_replication}).
The positive \method result concerns the complete operator under this joint shift in task, prompt, speakers, and acoustics.

\subsection{Parent parameterization affects source and transfer performance}

The operators in Tables~\ref{tab:headline_results} and~\ref{tab:factorial} edit block 20, the same site as DiReFT and LoReFT (Appendix Table~\ref{tab:baseline_contracts}).
For the matched parent analysis in Table~\ref{tab:training_tradeoff}, we select block 4 using development data.
TOFU operators are fitted on TOFU training authors, whereas SLUE operators are fitted on PopQA-20 and transferred unchanged.

\begin{table*}[t]
\centering
\tablestyle
\begin{tabular*}{\linewidth}{@{\extracolsep{\fill}}lrrrrr}
\toprule
& \multicolumn{2}{c}{TOFU} & \multicolumn{3}{c}{SLUE no-refit} \\
\cmidrule(lr){2-3}\cmidrule(lr){4-6}
\tablehead{Method} & \tablehead{RL} & \tablehead{EM} & \tablehead{Span} & \tablehead{RL} & \tablehead{EM} \\
\midrule
\directmethod (block 4) & 0.5588 & 0.1425 & \textbf{0.5000} & \textbf{0.5335} & \textbf{0.3750} \\
\expertmethod (block 4) & \textbf{0.5798} & \textbf{0.1888} & 0.4805 & 0.5275 & 0.3320 \\
\bottomrule
\end{tabular*}
\caption{\textbf{Parent parameterization affects the complete operator's source and transfer performance.}
\directmethod and \expertmethod combine the centered route field with the direct and expert parents, respectively.
Both rows edit decoder block 4.
TOFU reports RL and EM on 800 author-disjoint questions after TOFU fitting.
SLUE reports answer-span accuracy (Span), RL, and EM on 256 recorded questions after the corresponding PopQA-20-fitted operators transfer without refitting.}
\label{tab:training_tradeoff}
\end{table*}

With the centered route field included in both methods, \expertmethod scores higher on TOFU, while \directmethod has higher point estimates on SLUE.
On SLUE, \directmethod exceeds \expertmethod by +0.0430\,$\pm$\,0.0185 EM (bootstrap SE).
The point-estimate ranking of the two complete operators therefore differs between TOFU and SLUE.

\subsection{Scale, stability, and cross-backbone scope}

The full-scale PopQA results in Appendix Table~\ref{tab:full_popqa} preserve the gains at two operator-training budgets, and three Stage-B seeds reproduce both the PopQA-20 \directmethod and TOFU \expertmethod development results (Appendix Table~\ref{tab:seeds}).
The compact \factorizedmethod variant also improves held-out access on a separately acquired Kimi-Audio checkpoint, from 0.6321/0.0850 to 0.6451/0.1188 RL/EM.
This result supports portability of the compact variant to a second backbone under the same frozen speech-only constraint.
The primary Qwen \directmethod operator stores 0.46M learned values, edits one state, and uses one speech pass without a question transcript.
Appendix Table~\ref{tab:operational_contracts} compares it with speech LoRA, distillation, ASR, and full speech supervision under relaxed deployment constraints.

\section{Component and Depth Analysis}
\label{sec:ablations}

\paragraph{Shared parent and route field.}
We test each component's contribution under the same frozen one-state contract (Table~\ref{tab:factorial}).
Either component recovers substantial access, and their \directmethod composition gives the strongest RL across PopQA-20, PopQA-60, and TOFU.
Adding the route field to the shared parent improves RL and EM on both PopQA budgets, and adding the shared parent to the route field improves RL on both PopQA budgets (Appendix Table~\ref{tab:paired_intervals}).
On TOFU, the paired-bootstrap analysis does not resolve any of the four corresponding RL and EM differences.

\begin{table*}[t]
\centering
\tablestyle
\begin{tabular*}{\linewidth}{@{\extracolsep{\fill}}lrrrrrr}
\toprule
& \multicolumn{2}{c}{PopQA-20} & \multicolumn{2}{c}{PopQA-60} & \multicolumn{2}{c}{TOFU} \\
\cmidrule(lr){2-3}\cmidrule(lr){4-5}\cmidrule(lr){6-7}
\tablehead{Method} & \tablehead{RL} & \tablehead{EM} & \tablehead{RL} & \tablehead{EM} & \tablehead{RL} & \tablehead{EM} \\
\midrule
No Repair & 0.1787 & 0.1500 & 0.1787 & 0.1500 & 0.5003 & 0.1000 \\
+ Direct Parent & 0.3156 & 0.2986 & 0.3269 & 0.3087 & 0.5252 & 0.1138 \\
+ Route Field & 0.3212 & 0.3038 & 0.3379 & 0.3203 & 0.5265 & \textbf{0.1213} \\
\directmethod & \textbf{0.3353} & \textbf{0.3137} & \textbf{0.3493} & \textbf{0.3270} & \textbf{0.5310} & 0.1200 \\
\bottomrule
\end{tabular*}
\caption{\textbf{Component ablation of the shared access parent and centered route field.}
Rows beginning with ``+'' add one component to the frozen acquired checkpoint, and \directmethod combines both components.
PopQA-20, PopQA-60, and TOFU report RL and EM on their locked test splits.
All learned operators edit decoder block 20 and use development-matched edit norms.}
\label{tab:factorial}
\end{table*}

For the balanced PopQA \directmethod operator, before inference scaling, the shared parent vector has norm 31.48, the per-query route-field corrections have mean norm 8.69, and their mean cosine with the parent is 0.0104.
The mean centered route code over the operator-training set has norm below $2\times10^{-6}$.
This geometry is consistent with a dominant shared translation and a smaller, nearly orthogonal query-specific refinement.

\paragraph{Compact-parent depth.}
In the two-seed TOFU development sweep (Appendix~\ref{app:depth_characterization}), we hold the split, $K=1$ parent form, route-field coefficient, and target edit norm fixed across blocks 4 through 28.
Both seeds select block 4, and mean full-field RL declines from 0.5550 at block 4 to 0.4788 at block 28.
Within this compact variant, early blocks perform best, a pattern consistent with greater remaining decoder depth for integrating the edited prompt state.
The route field changes RL by at most 0.0068 at any block and seed, and the sign of its contribution varies across blocks and seeds.
The parent alone also peaks at block 4 in both seeds, so the depth trend comes from the parent.

\section{Discussion and Conclusion}
\label{sec:discussion}

The experiments show that text acquisition and speech access can diverge after an update.
\method implements speech-access repair as a detachable operator, combining a shared access parent with a smaller centered route field for query-specific variation.
This structure complements speech-pathway retraining when checkpoint immutability and one-pass speech inference matter.

On balanced synthetic speech, paired comparisons do not establish a clear \method advantage over SALSA or ReFT.
Operators fitted separately on full-scale PopQA differ in transfer to recorded SLUE questions with text passages, where ReFT and \method improve over No Repair while the tested SALSA variant regresses.

With the centered route field included, the expert parent gives higher block-4 TOFU scores and the direct parent gives higher SLUE point estimates.
The compact TOFU sweep favors early decoder blocks.

Knowledge can be acquired through the richest supervision route, and speech access can then be verified and repaired without modifying the acquired checkpoint.

\section*{Limitations}

We study short-answer factual access after text-only updates, primarily with Qwen2.5-Omni-7B.
Our PopQA and TOFU evaluations measure access to facts included in text acquisition, with disjoint operator-training, development, and test groups.
They therefore characterize transfer of the repair across acquired facts rather than acquisition of unseen knowledge.
The paired text--speech generation gap operationalizes access but does not localize every acoustic or internal cause.

Our PopQA and TOFU experiments use synthetic English speech from one fixed voice.
Recorded SLUE extends the evaluation to human speech, but jointly changes the task, prompt, speakers, and acoustics.
It also supplies a text passage, so this evaluation measures no-refit behavior transfer rather than closed-book recall of newly acquired facts from human recordings.
The transfer result concerns the complete fitted operator and does not isolate the centered route field's contribution.

Kimi-Audio provides a locked-test evaluation of the compact realization, while MiMo-Audio gives mixed development results.
The single-state intervention restricts adaptation capacity, and broader speech fine-tuning or an ASR cascade can attain higher accuracy when the deployment contract is relaxed.
Long-form generation, multilingual access, repeated checkpoint updates, and architecture-specific scaling remain future directions.

\section*{Ethical Considerations}

Our experiments concern access to factual information through spoken queries.
Improved access can make voice interfaces more useful, but an intervention can also change a previously correct answer or expose incorrect information stored by the model.
Aggregate benchmark gains do not establish reliability for consequential applications.
The English synthetic-speech and recorded-speech evaluations leave performance across languages, accents, and user groups incompletely characterized.
Use of the models and speech datasets remains subject to their respective licenses and usage conditions.

\paragraph{AI assistance.}
Generative-AI tools assisted with planning, experiment orchestration, code review, literature synthesis, analysis, and editing.
The authors verified the retained measurements against experiment artifacts and take responsibility for the paper's text, code, results, and claims.

\bibliography{references}

\appendix
\section{Training and Evaluation Details}
\label{app:reproduction}

We describe the data splits, model checkpoints, training procedures, and evaluation settings.

\subsection{Datasets and splits}

Table~\ref{tab:data_accounting} records the exact populations used by the reported experiments.
Before splitting PopQA, we group questions into connected components linked by shared subjects or normalized duplicate questions.
For TOFU, we group all questions about the same fictional author.
Each group belongs to a single operator-training, development, or test split.

\begin{table*}[t]
\centering
\tablestyle
\begin{tabular*}{\linewidth}{@{\extracolsep{\fill}}lrrrr}
\toprule
\tablehead{Setting} & \tablehead{Acquisition} & \tablehead{Operator train} & \tablehead{Development} & \tablehead{Test} \\
\midrule
Balanced PopQA & 1,000 & 600 & 199 & 201 \\
PopQA-20 & 14,267 & 2,853 & 2,848 & 2,847 \\
PopQA-60 & 14,267 & 8,560 & 2,848 & 2,847 \\
TOFU & 4,000 & 2,400 & 800 & 800 \\
SLUE-SQA-5 transfer & -- & 0 & 0 & 256 \\
\bottomrule
\end{tabular*}
\caption{Dataset and split accounting.
Acquisition is the text-update population seen by the acquired checkpoint.}
\label{tab:data_accounting}
\end{table*}

The PopQA-20 and PopQA-60 subsets are nested and are drawn only from the 8,572 operator-training rows.
The full acquired checkpoint saw every PopQA fact through text, including operator development and test facts.
This design measures generalization of the access operator across already acquired facts.
TOFU uses 120, 40, and 40 train, development, and test authors.
CosyVoice-300M-SFT renders every PopQA and TOFU speech question at 16 kHz with one fixed English female voice.
The same cached waveform is reused across methods for an item.
We evaluate on all 256 SLUE recordings from 50 speakers without fitting on SLUE.

\subsection{Acquired checkpoints and prompts}

Text acquisition updates all Qwen2.5-Omni-7B parameters for ten epochs on the corresponding text QA corpus.
It uses batch size one with four-step gradient accumulation, maximum sequence length 500, learning rate $2\times10^{-5}$, zero weight decay, a 0.03 warm-up ratio, and gradient-norm clipping at 1.
The balanced and full PopQA acquisitions are separate checkpoints, and TOFU uses its own acquired checkpoint.
Every access operator is fitted only after acquisition, and the acquired checkpoint remains frozen during operator fitting and evaluation.

Text questions use a plain question-answer prompt, while speech questions use the model's native audio wrapper with the same answer target.
The SLUE prompt contains a text passage followed by a recorded speech question under Qwen's standard multimodal system prompt.
No automatic transcript or text version of the spoken question is inserted for speech-only methods.
The five SLUE source datasets change both the linguistic task mixture and the prompt structure relative to short PopQA questions.

\subsection{State extraction and operator optimization}

The primary operator state is the final attended prompt position after decoder block 20.
The matched parent comparison and compact depth characterization use the blocks specified in their respective sections.
Qwen has hidden width $d=3{,}584$, and every low-rank operator uses rank $r=64$.
Table~\ref{tab:optimization} lists the trainable parameters and schedules for the \method variants.

\begin{table*}[t]
\centering
\tablestyle
\begin{tabular*}{\linewidth}{@{\extracolsep{\fill}}lP{3.3cm}rrr}
\toprule
\tablehead{Stage or method} & \tablehead{Trainable object} & \tablehead{Epochs} & \tablehead{Learning rate} & \tablehead{Seed} \\
\midrule
Paired geometry & $D$ and temporary readout & 1,000 steps & 0.001 & 5907 \\
\directmethod Stage A & Shared vector $g$ & 3 & 0.01 & 8307 \\
\directmethod Stage B & Local readout $U$ & 1 & 0.0001 & 6007 \\
\expertmethod Stage A & Router, 64 experts, and scale & 3 & 0.001/0.01/0.0001 & 8307 \\
\expertmethod Stage B & Local readout $U$ & 1 & 0.0001 & 6007 \\
\factorizedmethod Stage A & Direction $e$ and scalar $s$ & 3 & 0.01/0.0001 & 8307 \\
\factorizedmethod Stage B & Local readout $U$ & 1 & 0.0001 & 6007 \\
\bottomrule
\end{tabular*}
\caption{Main operator optimization.
Only the listed objects are trainable, and the acquired checkpoint remains frozen.}
\label{tab:optimization}
\end{table*}

Stage B uses batch size one, weight decay $10^{-4}$, gradient-norm clipping at 1, teacher temperature 1, and the top 32 teacher logits.
The item-level switch uses a 0.1-nat mean gold-answer log-probability advantage.
The sparse text distribution supervises the route field when text is better than the frozen selected parent by more than this threshold.
Otherwise, the frozen-parent distribution supplies the preservation target.
The route field is trained at unit output scale and deployed with $\beta=0.25$.

The stored learned-value counts are 3,584 for \globalfield, 458,752 for \routefield, 458,753 for \expertparent, 462,336 for \directmethod, 462,337 for \factorizedmethod, and 917,505 for \expertmethod.
Every realization edits one prompt-final state during the first generation step and requires no extra input under the speech-only contract.

\subsection{Frozen inference calibration}

We fix all intervention strengths before test evaluation.
Equal alpha values do not imply equal interventions across a norm-preserving vector and an affine field.
We therefore compare learned operators through realized mean edit norms rather than raw alphas.
The fixed \directmethod alphas are 0.447348 on PopQA-20, 0.429581 on PopQA-60, and 0.289593 on TOFU.
The corresponding \expertparent alphas are 0.869909, 0.760362, and 0.458232.
The PopQA-20 \directmethod alpha is reused unchanged for SLUE.
The block-4 \expertmethod alphas are 1.2469 on PopQA-20, 1.0672 on PopQA-60, and 0.826551 on TOFU.
Its PopQA-20 alpha is reused unchanged for SLUE.
The \globalfield alphas are 0.464829, 0.453870, and 0.294240.
For the matched Qwen operators, we calibrate to the same mean edit norm on development states.

For the compact cross-backbone characterization, the fixed Kimi-Audio and MiMo-Audio alphas are 0.601419 and 1.146747.
We calibrate these strengths on 2,280 cached Kimi operator-training states and 2,400 cached MiMo operator-training states without using answer scores.
The fixed target mean edit norms are 48.542453 for Kimi and 36.078838 for MiMo, taken from previously fitted affine operators on the respective backbones.

\subsection{Decoding, metrics, uncertainty, and failures}
\label{app:metrics}

Evaluation uses greedy decoding with at most 160 new tokens on PopQA and TOFU and at most 64 new tokens on SLUE.
ROUGE-L is the recall form of the longest-common-subsequence score against the reference answer.
Exact match lowercases text, strips punctuation and articles, normalizes whitespace, and then tests equality.
SLUE answer-span accuracy credits a generation when its normalized text contains the normalized short reference answer.
Retention divides the number of no-repair-correct items that remain exact under intervention by the total number of no-repair-correct items.

We report paired test differences with bootstrap standard errors (SEs), computed as the sample standard deviation of 10,000 group-bootstrap estimates with seed 3407.
Balanced PopQA resamples duplicate-question groups.
Full-scale PopQA resamples bound subject--question components for direct-parent and route-field contrasts, and retains the original relation-group resampling for expert-parent contrasts.
TOFU resamples authors, and SLUE resamples speakers.
Inferential statements and bolding use paired percentile-bootstrap bounds, computed separately from the SEs.
A difference is resolved when its 2.5th and 97.5th percentiles have the same nonzero sign.
These SEs quantify test-sample uncertainty, while Table~\ref{tab:seeds} reports sample standard deviations across training seeds.
All reported evaluations cover their stated sample counts with no failed generations.

\subsection{Software and hardware}

Qwen experiments use PyTorch 2.3.1, a Transformers 4.52.0 development snapshot, qwen-omni-utils 0.0.9, NumPy 2.0.2, and Python 3.9.23.
Model execution uses BF16 on NVIDIA A800-SXM4-80GB accelerators, while learned operator tensors and calibration statistics are retained in FP32.
Each training or generation run uses one accelerator.
The operational timing study in Section~\ref{app:operational} reports standalone end-to-end wall time and peak allocated CUDA memory for its matched runs.

\section{Baseline Implementations and Inference Inputs}
\label{app:baseline_replication}

We reproduce the central operation of each evaluated baseline on the factual-access datasets and select task-specific settings using development data.
Table~\ref{tab:baseline_contracts} summarizes the training data, intervention mechanisms, and additional inference inputs.

\begin{table*}[t]
\centering
\tablestyle
\begin{tabular*}{\linewidth}{@{\extracolsep{\fill}}P{4.4cm}P{3.2cm}P{7.2cm}}
\toprule
\tablehead{Method} & \tablehead{Training data} & \tablehead{Mechanism and inference inputs} \\
\midrule
DiReFT~\citep{wu2024reft} & 600 balanced, 2,853 PopQA-20, or 2,400 TOFU rows & Rank-64 low-rank block-20 edit with a development-selected source dose \\
LoReFT~\citep{wu2024reft} & 600 balanced, 2,853 PopQA-20, or 2,400 TOFU rows & Rank-64 orthogonal low-rank block-20 edit with a development-selected source dose \\
L2S~\citep{parekh2025learning} & 600 balanced, 2,853 PopQA-20, or 2,400 TOFU paired states & Context-conditioned predictor with a 100-dimensional bottleneck, with intervention strength calibrated by matching edit norms on source states without answer scores \\
SALSA~\citep{yegorova2026salsa} & 600 balanced, 2,853 PopQA-20, or 2,400 TOFU rows & One norm-preserving steering activation vector with a development-selected or source-state norm-matched dose \\
Angle~\citep{xiang2025modalitygap} & Paired speech and text development states only & No fitted operator, a paired text-question donor, and the development-selected all-coordinate angle rule \\
Length~\citep{xiang2025modalitygap} & Paired speech and text development states only & No fitted operator, a paired text-question donor, and the development-selected bottom-three length rule \\
\bottomrule
\end{tabular*}
\caption{Baseline training data and inference inputs.
Balanced PopQA, full PopQA-20, and TOFU use separately fitted artifacts.
Angle and Length are donor-assisted controls because they receive the paired text question at inference.}
\label{tab:baseline_contracts}
\end{table*}

DiReFT and LoReFT use one answer cross-entropy epoch with learning rate $10^{-3}$, batch size one, and gradient-norm clipping at 1.
Their full PopQA-20 source-development doses are 0.073351 and 0.212747.
These doses are fixed before the operators are applied unchanged to all 256 SLUE items.

L2S fits paired text and speech states for 100 epochs with batch size 64, learning rate $5\times10^{-5}$, a maximum cosine weight of 0.1, and an internal 8.33\% validation fraction.
Its balanced and full PopQA-20 score-blind alphas are 0.415363 and 0.463409.
The full artifact is applied unchanged to SLUE.

The balanced-PopQA SALSA artifact is trained for three epochs at learning rate $5\times10^{-4}$ and uses development-selected alpha 3.600604.
That artifact and alpha produce only the 201-item balanced-PopQA result in Table~\ref{tab:headline_results}.
The SLUE row instead uses a full PopQA-20 artifact fitted on 2,853 operator-training examples.
Score-blind calibration on those source speech states matches \method's source mean edit norm of 25.448338 and selects alpha 2.012277.
The frozen SALSA artifact then realizes mean edit norm 25.1334 on SLUE.

In a separate control, we recalibrate SALSA using unlabeled SLUE inputs to match \method's mean edit norm on the target inputs.
It selects SALSA alpha 1.886724 and matches \method's realized SLUE mean edit norm to within 0.031\%.
SALSA remains below its paired No Repair replay, with RL 0.3759 versus 0.4437 and a difference of \textminus0.0678\,$\pm$\,0.0272 (bootstrap SE).
Relative to source-calibrated SALSA, the dose reduction changes RL by only +0.0029\,$\pm$\,0.0088 (bootstrap SE).
SALSA's negative RL transfer persists after matching the mean edit norm on SLUE inputs.

Angle and Length use the paired text-question state at inference.
On balanced PopQA development data, we select Angle over all coordinates and Length over the bottom three coordinates by RL with EM as the tie-breaker.
The selected rule is then reused for balanced PopQA test and recorded SLUE.
We apply the same development-only rule selection to TOFU using its author-disjoint split.

\section{Statistical Evidence for Main Comparisons}
\label{app:contrasts}

This section reports paired differences with bootstrap SEs for the comparisons that carry the main empirical argument.
Bold estimates satisfy the paired-bootstrap criterion in Appendix~\ref{app:metrics}.
Each method-minus-baseline difference uses the No Repair outputs generated alongside that method unless a method-to-method contrast is stated explicitly.

\subsection{Balanced source-domain comparisons}

Table~\ref{tab:published_intervals} reports paired group-bootstrap comparisons on the balanced PopQA test set.
We compute SEs from paired per-item score differences between \directmethod and each comparator.

\begin{table*}[t]
\centering
\tablestyle
\begin{tabular*}{\linewidth}{@{\extracolsep{\fill}}lrr}
\toprule
\tablehead{Comparator} & \tablehead{$\Delta$ RL $\pm$ SE} & \tablehead{$\Delta$ EM $\pm$ SE} \\
\midrule
DiReFT & +0.03731\,$\pm$\,0.02074 & +0.03980\,$\pm$\,0.02122 \\
LoReFT & +0.03234\,$\pm$\,0.02048 & +0.02985\,$\pm$\,0.02120 \\
SALSA & +0.01990\,$\pm$\,0.01604 & +0.01990\,$\pm$\,0.01719 \\
\expertparent & \textminus0.00249\,$\pm$\,0.01229 & \textminus0.00498\,$\pm$\,0.01325 \\
\bottomrule
\end{tabular*}
\caption{\textbf{\directmethod minus matched single-state comparators on balanced PopQA.}
We estimate SEs from 10,000 bootstrap resamples of duplicate-question groups.}
\label{tab:published_intervals}
\end{table*}

The paired-bootstrap analysis does not resolve the RL differences.

\subsection{No-refit recorded-speech transfer}

Table~\ref{tab:slue_paired_baselines} reports every speech-only method relative to the No Repair outputs generated alongside it.
The \method/ReFT and L2S/SALSA evaluations generated these outputs separately.
The archived \directmethod and SALSA configurations specify the same acquired checkpoint, question manifest, system prompt, and generation limit, but their No Repair outputs differ on some items.
The source of these differences has not been isolated, so each method's paired improvement uses its own generated baseline rather than substituting the baseline from another run.

\begin{table*}[t]
\centering
\tablestyle
\begin{tabular*}{\linewidth}{@{\extracolsep{\fill}}lrrr}
\toprule
\tablehead{Method} & \tablehead{$\Delta$ Span $\pm$ SE} & \tablehead{$\Delta$ RL $\pm$ SE} & \tablehead{$\Delta$ EM $\pm$ SE} \\
\midrule
DiReFT & \textbf{+0.05469\,$\pm$\,0.01720} & \textbf{+0.05456\,$\pm$\,0.01737} & \textbf{+0.03906\,$\pm$\,0.01388} \\
LoReFT & \textbf{+0.07031\,$\pm$\,0.02386} & \textbf{+0.07734\,$\pm$\,0.02225} & \textbf{+0.07031\,$\pm$\,0.01610} \\
L2S & \textminus0.01953\,$\pm$\,0.02197 & \textminus0.01758\,$\pm$\,0.02083 & +0.02344\,$\pm$\,0.01828 \\
SALSA & \textbf{\textminus0.08984\,$\pm$\,0.03081} & \textbf{\textminus0.07070\,$\pm$\,0.02643} & +0.02344\,$\pm$\,0.02400 \\
\directmethod & \textbf{+0.08594\,$\pm$\,0.02507} & \textbf{+0.09408\,$\pm$\,0.02450} & \textbf{+0.06641\,$\pm$\,0.01944} \\
\bottomrule
\end{tabular*}
\caption{\textbf{No-refit SLUE method minus its paired No Repair replay.}
We estimate SEs from 10,000 bootstrap resamples of the 50 speakers.}
\label{tab:slue_paired_baselines}
\end{table*}

The paired-bootstrap analysis supports positive differences for the complete \directmethod operator on all three recorded-speech metrics.
DiReFT and LoReFT also improve over their paired baselines, while SALSA has a resolved negative RL difference.
The bootstrap intervals for L2S's differences from No Repair include zero on all three metrics.

\subsection{Full-scale component contrasts}

Table~\ref{tab:paired_intervals} reports the paired differences and SEs behind the component analysis.
Positive values favor \directmethod.

\begin{table*}[t]
\centering
\tablestyle
\begin{tabular*}{\linewidth}{@{\extracolsep{\fill}}llrr}
\toprule
\tablehead{Setting} & \tablehead{Comparator} & \tablehead{$\Delta$ RL $\pm$ SE} & \tablehead{$\Delta$ EM $\pm$ SE} \\
\midrule
PopQA-20 & \globalfield & \textbf{+0.01974\,$\pm$\,0.00425} & \textbf{+0.01510\,$\pm$\,0.00442} \\
 & \routefield & \textbf{+0.01412\,$\pm$\,0.00484} & \textbf{+0.00983\,$\pm$\,0.00495} \\
 & \expertparent & +0.00083\,$\pm$\,0.00363 & \textminus0.00176\,$\pm$\,0.00381 \\
\midrule
PopQA-60 & \globalfield & \textbf{+0.02232\,$\pm$\,0.00494} & \textbf{+0.01826\,$\pm$\,0.00511} \\
 & \routefield & \textbf{+0.01135\,$\pm$\,0.00507} & +0.00667\,$\pm$\,0.00521 \\
 & \expertparent & +0.00226\,$\pm$\,0.00494 & \textbf{+0.01054\,$\pm$\,0.00600} \\
\midrule
TOFU & \globalfield & +0.00577\,$\pm$\,0.00364 & +0.00625\,$\pm$\,0.00404 \\
 & \routefield & +0.00454\,$\pm$\,0.00530 & \textminus0.00125\,$\pm$\,0.00448 \\
 & \expertparent & +0.00369\,$\pm$\,0.00561 & \textminus0.00125\,$\pm$\,0.00412 \\
\midrule
SLUE & \routefield & \textbf{+0.06283\,$\pm$\,0.02143} & +0.03516\,$\pm$\,0.01805 \\
 & \expertparent & +0.02591\,$\pm$\,0.01585 & +0.03516\,$\pm$\,0.01915 \\
\bottomrule
\end{tabular*}
\caption{\textbf{\directmethod minus full-scale component and parent comparators.}
SEs use the group-bootstrap procedures in Appendix~\ref{app:metrics}.}
\label{tab:paired_intervals}
\end{table*}

For SLUE answer-span accuracy, \directmethod minus \routefield is +0.06250\,$\pm$\,0.02163 (bootstrap SE).
The corresponding difference from \expertparent is +0.03516\,$\pm$\,0.01835 (bootstrap SE).

\subsection{Block-4 repair comparisons}

Table~\ref{tab:l4_transfer_intervals} compares block-4 \directmethod with the block-4 \expertmethod and the block-20 expert parent on the same SLUE questions.
Positive values favor \directmethod.

\begin{table*}[t]
\centering
\tablestyle
\begin{tabular*}{\linewidth}{@{\extracolsep{\fill}}lrrr}
\toprule
\tablehead{Comparator} & \tablehead{$\Delta$ Span $\pm$ SE} & \tablehead{$\Delta$ RL $\pm$ SE} & \tablehead{$\Delta$ EM $\pm$ SE} \\
\midrule
\expertparent (block 20) & +0.0430\,$\pm$\,0.0255 & +0.0218\,$\pm$\,0.0216 & \textbf{+0.1094\,$\pm$\,0.0251} \\
\expertmethod (block 4) & +0.0195\,$\pm$\,0.0282 & +0.0060\,$\pm$\,0.0266 & \textbf{+0.0430\,$\pm$\,0.0185} \\
\bottomrule
\end{tabular*}
\caption{\textbf{Block-4 \directmethod minus its comparators on no-refit SLUE.}
We estimate SEs from 10,000 bootstrap resamples of the 50 speakers.
The expert-parent row uses a different intervention block.}
\label{tab:l4_transfer_intervals}
\end{table*}

The paired-bootstrap analysis supports a resolved exact-match advantage for \directmethod on this joint shift, with positive point estimates in answer span and RL.

\section{Full Results and Characterization}
\label{app:controls}

\subsection{Synthesized-question content retention}

To test whether synthetic-speech states preserve question-specific content, we analyze the 600 paired text and CosyVoice speech prompt-final states captured for \method fitting at block 20.
After subtracting the route-specific population mean, speech-to-text exact-pair retrieval reaches 0.5033 top-1, 0.6967 top-5, and 0.5979 mean reciprocal rank over all 600 candidates.
Global top-1 chance is 0.0017.
When candidates are restricted to the same PopQA relation, top-1 remains 0.5033 while item-averaged chance is 0.0267 with a median of 52 candidates.
The reverse text-to-speech direction gives 0.5183 global and 0.5200 within-relation top-1.
These results show that the internal speech state preserves substantial item-specific question identity rather than only the relation template.
They rule out total loss of question identity as a sufficient explanation but do not isolate the cause of the remaining generation gap.

\subsection{Full-scale PopQA matrix}

\begin{table*}[t]
\centering
\tablestyle
\begin{tabular*}{\linewidth}{@{\extracolsep{\fill}}lrrrrr}
\toprule
& & \multicolumn{2}{c}{PopQA-20} & \multicolumn{2}{c}{PopQA-60} \\
\cmidrule(lr){3-4}\cmidrule(lr){5-6}
\tablehead{Method} & \tablehead{Block} & \tablehead{RL} & \tablehead{EM} & \tablehead{RL} & \tablehead{EM} \\
\midrule
No Repair & -- & 0.1787 & 0.1500 & 0.1787 & 0.1500 \\
\globalfield & 20 & 0.3156 & 0.2986 & 0.3269 & 0.3087 \\
\routefield & 20 & 0.3212 & 0.3038 & 0.3379 & 0.3203 \\
\expertparent & 20 & 0.3345 & 0.3154 & 0.3470 & 0.3165 \\
\directmethod & 20 & 0.3353 & 0.3137 & 0.3493 & 0.3270 \\
\midrule
\expertmethod & 4 & \textbf{0.3727} & \textbf{0.3562} & \textbf{0.3871} & \textbf{0.3709} \\
\bottomrule
\end{tabular*}
\caption{\textbf{Full-scale PopQA factual access.}
PopQA-20 and PopQA-60 use operator-training budgets of approximately 20\% and 60\% of the full acquisition population and share the same locked test set.
Block is shown because only same-block rows are controlled architectural comparisons.}
\label{tab:full_popqa}
\end{table*}

We evaluate block-4 \expertmethod on 2,847 test questions.
A block-4 parent-only PopQA test result is unavailable, so this row does not provide a same-layer component ablation.

\subsection{Stage-B seed stability}

Table~\ref{tab:seeds} summarizes development performance across three Stage-B training runs for PopQA-20 and TOFU.

\begin{table*}[t]
\centering
\tablestyle
\begin{tabular*}{\linewidth}{@{\extracolsep{\fill}}lrrr}
\toprule
\tablehead{Method and setting} & \tablehead{Development RL} & \tablehead{Development EM} & \tablehead{Mean edit norm} \\
\midrule
\directmethod, PopQA-20 & 0.3462\,$\pm$\,0.0018 & 0.3246\,$\pm$\,0.0017 & 25.469\,$\pm$\,0.020 \\
\midrule
\expertmethod, TOFU & 0.5619\,$\pm$\,0.0019 & 0.1571\,$\pm$\,0.0031 & $\approx$ 19.232 \\
\bottomrule
\end{tabular*}
\caption{\textbf{Stage-B stability across three training seeds.}
RL and EM are reported as mean $\pm$ sample standard deviation.}
\label{tab:seeds}
\end{table*}

PopQA-20 development scores are stable across the three Stage-B optimizer seeds at a fixed intervention strength.
For TOFU, each seed uses an intervention strength calibrated to the same target mean edit norm.

\subsection{Compact-parent depth characterization}
\label{app:depth_characterization}

We evaluate \factorizedmethod across decoder blocks using two training seeds and 800 TOFU development items, as summarized in Section~\ref{sec:ablations}.
The split, $K=1$ parent parameterization, local coefficient, and target edit norm are fixed across blocks 4 through 28.

Mean full-field RL declines from 0.5550 at block 4 to 0.4788 at block 28.
The route field changes RL by at most 0.0068 at any block and seed, its two-seed average never exceeds 0.0025, and its sign varies across blocks and seeds.
The parent alone also peaks at block 4 in both seeds, so the route field does not determine the selected site.

\subsection{Cross-backbone characterization}
\label{app:cross_model}

Table~\ref{tab:cross_backbone_factorized} compares the compact parent-plus-route-field variant across speech-language backbones.
Each backbone uses its own frozen text-acquired checkpoint, with intervention strengths calibrated on the states specified in Appendix~\ref{app:reproduction}.
The local coefficient $\beta=0.25$ is transferred without backbone-specific retuning.

\begin{table*}[t]
\centering
\tablestyle
\begin{tabular*}{\linewidth}{@{\extracolsep{\fill}}llrrrrrrr}
\toprule
& & & \multicolumn{2}{c}{No Repair} & \multicolumn{2}{c}{Parent} & \multicolumn{2}{c}{Full field} \\
\cmidrule(lr){4-5}\cmidrule(lr){6-7}\cmidrule(lr){8-9}
\tablehead{Backbone} & \tablehead{Split} & \tablehead{Block} & \tablehead{RL} & \tablehead{EM} & \tablehead{RL} & \tablehead{EM} & \tablehead{RL} & \tablehead{EM} \\
\midrule
Qwen2.5-Omni-7B & Development & 4 & 0.4695 & 0.0713 & 0.5575 & 0.1613 & 0.5585 & 0.1563 \\
Kimi-Audio & Development & 20 & 0.5655 & 0.0875 & 0.5973 & 0.1338 & 0.5960 & 0.1350 \\
Kimi-Audio & Test & 20 & 0.6321 & 0.0850 & -- & -- & \textbf{0.6451} & \textbf{0.1188} \\
MiMo-Audio & Development & 25$\to$36 & 0.7175 & 0.2450 & 0.7113 & 0.2563 & 0.7130 & 0.2575 \\
\bottomrule
\end{tabular*}
\caption{\textbf{\factorizedmethod characterization across speech-language backbones on TOFU.}
Each reported score is computed over 800 questions.
Development rows show No Repair, the parent, and the full field for each backbone.
For Kimi, we also report test results after model selection on development data.}
\label{tab:cross_backbone_factorized}
\end{table*}

We evaluate the selected Kimi operator on a separate test set.
Its full field improves test RL and EM from 0.6321/0.0850 to 0.6451/0.1188 and raises 233 item-level ROUGE-L scores.
A second training seed on Kimi-Audio reaches development RL/EM of 0.5992/0.1325 against the same 0.5655/0.0875 baseline.
MiMo is mixed, with a 0.0125 EM increase and a 0.0044 RL decrease relative to No Repair.

\subsection{Geometric and capacity controls}

To assess query-dependent geometry and model capacity, we compare a constant route mean, an answer-free low-rank route map, and a high-capacity kernel regressor on balanced PopQA (Table~\ref{tab:geometric_controls}).
All nonzero edits operate at block 20 and use the same speech-only deployment interface.

\begin{table*}[t]
\centering
\tablestyle
\begin{tabular*}{\linewidth}{@{\extracolsep{\fill}}lrrr}
\toprule
\tablehead{Method} & \tablehead{Learned values} & \tablehead{RL} & \tablehead{EM} \\
\midrule
No Repair & 0 & 0.2736 & 0.2189 \\
Mean route shift & 3,584 & 0.5323 & 0.5224 \\
Representation-only route map & 458,752 & 0.5808 & 0.5672 \\
Residual KRR~\citep{saunders1998ridge} & 4,300,800 & 0.5808 & 0.5672 \\
\directmethod & 462,336 & \textbf{0.6318} & \textbf{0.6219} \\
\bottomrule
\end{tabular*}
\caption{\textbf{Representation controls on the 201-item balanced PopQA test.}
The mean edit norms are 22.51 for the controls and 22.62 for \directmethod.}
\label{tab:geometric_controls}
\end{table*}

The route map achieves higher RL and EM than the mean shift on balanced PopQA.
Residual KRR stores 9.4 times more learned values than the route map but does not improve its test scores.
The remaining \directmethod gap is consistent with decoder-facing training contributing beyond representation-only mapping.

\section{Comparison with Speech Fine-Tuning and ASR}
\label{app:operational}

We compare external repair operators with alternatives that update model parameters or add an ASR pass.

\subsection{Matched-data direct speech fine-tuning}

Table~\ref{tab:affirm_speech_sft_summary} summarizes a development comparison between a decoder-trained route map and direct Speech LoRA.
The four data budgets use nested training subsets and the same 199-item development set.
We evaluate this sweep only on development data.
The route map first fits rank-64 paired-state geometry for 1,000 steps, then updates its readout for one epoch at learning rate 0.001 while retaining the empirical mean route difference.
Its selective distillation uses the text route as the correction teacher and the unmodified speech route as the preservation teacher.
This variant has no separately trained access parent, so its resource measurements describe the route map rather than the staged \method procedure in Section~\ref{sec:method}.
We calibrate its intervention strength without answer scores to the same mean edit norm at every budget.
Speech LoRA uses rank-16 adapters in all 28 decoder blocks and a common five-epoch cosine schedule.
For each Speech LoRA budget, the table reports the development checkpoint selected by RL with EM as the tie-breaker.

\begin{table*}[t]
\centering
\tablestyle
\begin{tabular*}{\linewidth}{@{\extracolsep{\fill}}lrrrrrrr}
\toprule
\tablehead{Method} & \tablehead{Unique pairs} & \tablehead{Selected epoch} & \tablehead{RL} & \tablehead{EM} & \tablehead{Time (s)} & \tablehead{Peak GiB} & \tablehead{Learned values} \\
\midrule
Route map & 75 & -- & 0.5343 & 0.5226 & 123.0 & 21.31 & 0.46M \\
Speech LoRA & 75 & 3 & 0.5419 & 0.5276 & 267.4 & 17.61 & 40.37M \\
\addlinespace
Route map & 150 & -- & 0.5720 & 0.5628 & 202.6 & 21.42 & 0.46M \\
Speech LoRA & 150 & 2 & 0.5972 & 0.5829 & 339.9 & 17.69 & 40.37M \\
\addlinespace
Route map & 300 & -- & 0.5494 & 0.5327 & 468.7 & 21.42 & 0.46M \\
Speech LoRA & 300 & 3 & 0.6323 & 0.6231 & 575.5 & 17.69 & 40.37M \\
\addlinespace
Route map & 600 & -- & 0.5771 & 0.5628 & 951.4 & 21.42 & 0.46M \\
Speech LoRA & 600 & 2 & 0.6323 & 0.6281 & 913.8 & 17.68 & 40.37M \\
\bottomrule
\end{tabular*}
\caption{\textbf{Matched-data development comparison of a decoder-trained route map and Speech LoRA.}
The route map uses the empirical mean route difference in place of \method's trained access parent and the unmodified speech output as its preservation target.
Its learned-value count includes both projection matrices and the empirical mean route difference.
Time is standalone end-to-end training time in seconds and excludes development generation.
Speech LoRA checkpoint times are reconstructed from checkpoint timestamps and stored full-run duration, so they are approximate.}
\label{tab:affirm_speech_sft_summary}
\end{table*}

After selecting among five development checkpoints, Speech LoRA is at least as accurate at every unique-pair budget.
At 150 pairs and approximately matched wall time, the route map reaches 0.5720/0.5628 RL/EM in 202.6 seconds, while the first Speech LoRA epoch reaches 0.5695/0.5578 in 197.8 seconds.
The route map stores fewer learned values, while Speech LoRA uses less peak training memory at every budget and trains faster at the largest budget.

\subsection{Test accuracy and deployment requirements}

Table~\ref{tab:operational_contracts} compares test accuracy and deployment requirements on balanced PopQA, PopQA-60, and TOFU.

\begin{table*}[t]
\centering
\tablestyle
\begin{tabular*}{\linewidth}{@{\extracolsep{\fill}}P{2.4cm}P{3.0cm}rrP{3.2cm}rr}
\toprule
\tablehead{Setting} & \tablehead{Method} & \tablehead{Pairs} & \tablehead{Trainable} & \tablehead{Deployed change} & \tablehead{RL} & \tablehead{EM} \\
\midrule
Balanced PopQA & No Repair & 0 & 0 & None & 0.2736 & 0.2189 \\
& \directmethod & 600 & 0.46M & One-state external operator & 0.6318 & 0.6219 \\
& Speech LoRA & 600 & 40.37M & Internal adapters in 28 blocks & 0.6443 & 0.6368 \\
& ASR cascade & 0 & 0 & External ASR and a second text pass & 0.6990 & 0.6915 \\
& Full speech supervision & 1,000 & Full model & Full-model update & 0.9801 & 0.9801 \\
\midrule
PopQA-60 & No Repair & 0 & 0 & None & 0.1787 & 0.1500 \\
& \directmethod & 8,560 & 0.46M & One-state external operator & 0.3493 & 0.3270 \\
& Speech LoRA & 8,560 & 40.37M & Internal adapters in 28 blocks & 0.4962 & 0.4830 \\
& S2T-KD & 8,560 & 40.37M & Internal adapters in 28 blocks & 0.4943 & 0.4805 \\
\midrule
TOFU & No Repair & 0 & 0 & None & 0.5003 & 0.1000 \\
& \directmethod & 2,400 & 0.46M & One-state external operator & 0.5310 & 0.1200 \\
& Speech LoRA & 2,400 & 40.37M & Internal adapters in 28 blocks & 0.6509 & 0.2513 \\
& S2T-KD & 2,400 & 40.37M & Internal adapters in 28 blocks & 0.6504 & 0.2275 \\
\bottomrule
\end{tabular*}
\caption{\textbf{Accuracy and mutability across frozen and relaxed deployment contracts.}
All scores are computed on the full test splits.
Speech LoRA and S2T-KD insert rank-16 adapters throughout all 28 decoder blocks, and S2T-KD adds a temperature-2 forward-KL term with weight 0.5 toward the text-route output distribution.
The ASR cascade transcribes each spoken question with Whisper-large-v3 and answers the transcript with the acquired checkpoint.}
\label{tab:operational_contracts}
\end{table*}

Broader adaptation and extra inference resources can reach higher accuracy than the detachable one-state operator.

\end{document}